%% file: main.tex
\documentclass[runningheads]{llncs}
\usepackage[T1]{fontenc}
\usepackage{graphicx,verbatim}

\usepackage{amsmath}
\usepackage{amssymb}
\usepackage{xspace}
\usepackage[table,xcdraw,dvipsnames]{xcolor}
\usepackage{hyperref}

\newcommand{\figref}[1]{Fig.~\ref{#1}}
\newcommand{\ie}{\emph{i.e.}}
\newcommand{\eg}{\emph{e.g.}}

\begin{document}
\title{Fetuses Made Simple: Modeling and Tracking of Fetal Shape and Pose}

\author{Yingcheng Liu\inst{1} \and
Peiqi Wang\inst{1} \and
Sebastian Diaz\inst{1} \and
Esra Abaci Turk\inst{2} \and
Benjamin Billot\inst{3} \and
P. Ellen Grant\inst{2} \and
Polina Golland\inst{1}}

\authorrunning{Y. Liu et al.}

\institute{Computer Science and Artificial Intelligence Lab, MIT, Cambridge, USA \and
Boston Children’s Hospital and Harvard Medical School, Boston, USA \and
Inria, Epione team, Sophia-Antipolis, France \\
\email{liuyc@mit.edu}}

\maketitle              %

\input{texts/abstract}

\input{texts/introduction}

\input{texts/method}

\input{texts/experiment}

\input{texts/conclusion}

\bibliographystyle{splncs04}
\bibliography{bibliography}

\clearpage
\appendix
\renewcommand{\thesection}{\Alph{section}}
\setcounter{section}{0}

\begin{center}
\LARGE \textbf{Appendices}
\end{center}

\input{texts/appendix.tex}

\end{document}

%% file: texts/abstract.tex
\begin{abstract}

Analyzing fetal body motion and shape is paramount in prenatal diagnostics and monitoring. 
Existing methods for fetal MRI analysis mainly rely on anatomical keypoints or volumetric body segmentations.
Keypoints simplify body structure to facilitate motion analysis, but may ignore important details of full-body shape. 
Body segmentations capture complete shape information but complicate temporal analysis due to large non-local fetal movements. 
To address these limitations, we construct a 3D articulated statistical fetal body model based on the Skinned Multi-Person Linear Model (SMPL). 
Our algorithm iteratively estimates body pose in the image space and body shape in the canonical pose space. 
This approach improves robustness to MRI motion artifacts and intensity distortions, and reduces the impact of incomplete surface observations due to challenging fetal poses. 
We train our model on segmentations and keypoints derived from $19,816$ MRI volumes across $53$ subjects.
Our model captures body shape and motion across time series and provides intuitive visualization.
Furthermore, it enables automated anthropometric measurements traditionally difficult to obtain from segmentations and keypoints.
When tested on unseen fetal body shapes, our method yields a surface alignment error of $3.2$~mm for $3$~mm MRI voxel size.
To our knowledge, this represents the first 3D articulated statistical fetal body model, paving the way for enhanced fetal motion and shape analysis in prenatal diagnostics.
The code is available at \url{https://github.com/MedicalVisionGroup/fetal-smpl}.

\keywords{Fetal MRI \and Pose estimation \and Shape estimation \and 3D body model \and SMPL (Skinned Multi-Person Linear Model)}

\end{abstract}

%% file: texts/introduction.tex
\section{Introduction}
\label{sec:intro}

Analyzing fetal body motion and shape is crucial in prenatal diagnosis and monitoring.
For example, fetal movement is related to the function of the fetal central nervous system~\cite{vasung2023cross}, and anatomical growth patterns provide key biomarkers of fetal health~\cite{snijders1994fetal}.
In the current clinical workflow, most assessments are performed visually with 2D ultrasound. 
When ultrasound findings are inconclusive, 3D MRI is an established solution that visualizes fetal anatomy fully in 3D.
Recent advances in machine learning have enabled automatic fetal MRI analysis~\cite{vasung2023cross,xu2019fetal,xu2021motion,billot2024se,sajapala20174d}. 
These methods typically model the fetal body as a set of anatomical landmarks~\cite{vasung2023cross,xu2019fetal,xu2021motion} or rigid body parts~\cite{billot2024se,sajapala20174d}.
Keypoint-based approaches facilitate motion analysis by representing the body as a sparse ``stick figure,'' but this simplification may ignore important full-body shape information.
Alternatively, volumetric segmentation captures the full body shape~\cite{uus2024automated,sadhwani2022fetal}.
However, motion analysis with segmentation is challenging because the fetal body undergoes large deformations. 
Existing deformable registration methods often assume small local deformations and struggle to handle the wide variability in pose and large non-local deformations of the fetal body~\cite{chi2023dynamic}. 

In this paper, we aim to overcome these limitations by building a surface-based kinematic fetal body model using segmentation and keypoints from MRI time series.
Our model is based on Skinned Multi-Person Linear Model (SMPL)~\cite{loper2023smpl}.
We represent the fetal body as a collection of connected rigid parts parameterized by a kinematic tree.
The kinematic tree enables us to align fetal body into a standardized pose, where we define a canonical space and subject-specific body shape.
This approach decouples the representations of the pose-induced deformation and intrinsic body shape.
Once trained and aligned to images, our model facilitates the extraction of surface correspondence between frames for body motion analysis.

\input{figtexs/fetal_rave.tex}

Building a fetal SMPL model using MRI is challenging.
In adult SMPL model training, the correspondence between the model and data surface is known~\cite{loper2023smpl}. 
When the correspondence is not available, existing methods often require the subject to lie in a standard pose~\cite{alldieck2018video}.
The body shape is then estimated by minimizing the Chamfer distance between the model and data surfaces~\cite{alldieck2018video}.
In our application, placing the fetuses in a shared standard pose during the MRI imaging procedure is impossible.
Furthermore, our images are low-resolution and frequently contain noise from motion artifacts or maternal breathing.
Surface observations are also often incomplete.
Moreover, body parts (\eg, legs and abdomen) frequently touch, making it difficult to extract the skin surface from volumetric segmentations.

To tackle these challenges, we iteratively estimate body shape and pose using a coordinate descent algorithm.
First, we estimate pose using an initial body shape and ``unpose'' the observed body surfaces in all frames into the common canonical space.
We then reconstruct the subject-specific canonical shape using all the unposed body surfaces.
This enables us to aggregate shape information across frames and mitigate the impact of noisy, incomplete surface observations in individual frames.
Subsequently, the pose estimate is refined using the updated shape estimates.
We repeat these steps until shape and pose estimates converge.
This approach is inspired by prior work on adult and newborn body modeling~\cite{alldieck2018video,hesse2019learning,hesse2018learning}, adapted specifically for fetal MRI.
Additionally, we improve the method by deriving exact coordinate transformations with inverse kinematics,
which supports accurate mapping of the segmentation vertices to the canonical space.

We train fetal SMPL using segmentations and keypoints derived from $19,816$ volumes of Echo Planar Imaging (EPI) MRI data across $53$ subjects.
\figref{fig:fetal_rave} illustrates the dynamics of our model and an overview of the registered model.
The model captures diverse fetal shape and pose, providing new tools for fetal MRI analysis.
Experiments show that our model explains unseen shapes with a surface alignment error of $3.2$~mm in volumes with $3$~mm isotropic voxels.
The resulting anthropometric measurements illustrate potential applications for developmental abnormality detection.

%% file: figtexs/fetal_rave.tex
\begin{figure}[tp]
    \centering
    \includegraphics[width=\linewidth]{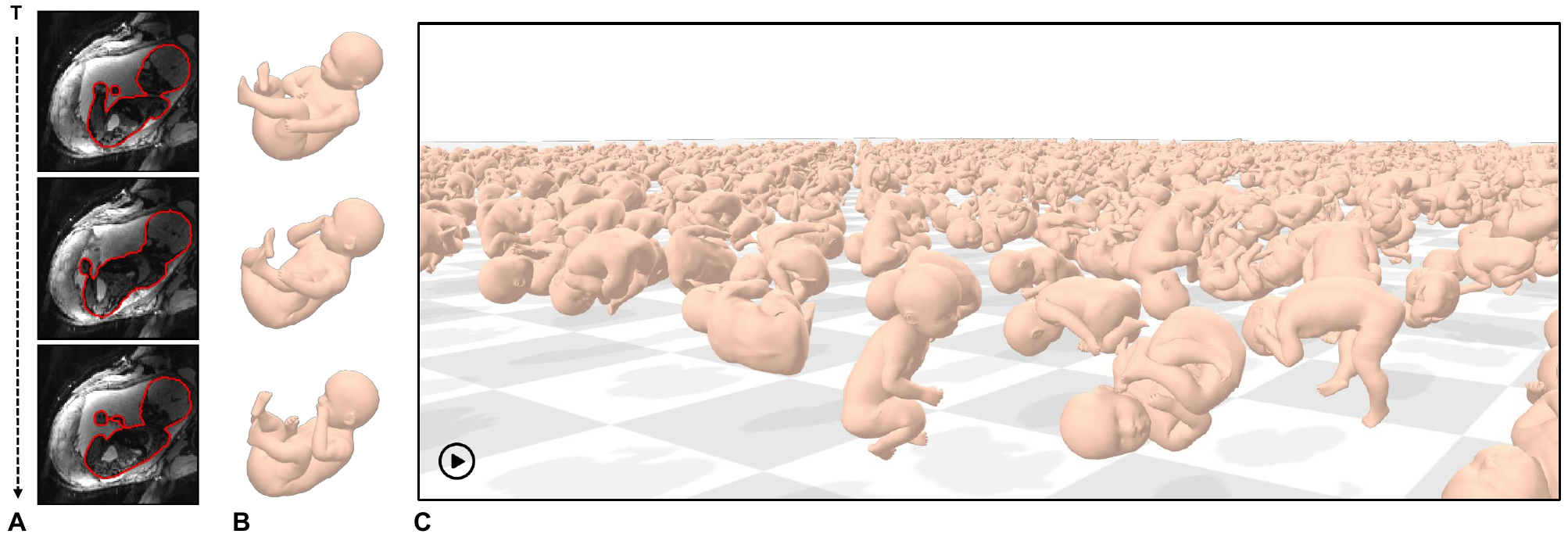}
    \caption{
        \textbf{Fetal rave.}
        We learn a 3D articulated statistical fetal body shape model from $20$k volumes of MRI time series in $53$ subjects.
        (\textbf{A}): 3D MRI time series with fetal body segmentations.
        (\textbf{B}): Model aligned to volumetric frames in \textbf{A}.
        (\textbf{C}): Overview of our model aligned to all volumes in our dataset. 
    }
    \label{fig:fetal_rave}
\end{figure}

%% file: texts/method.tex
\section{Method}
\label{sec:method}

\subsection{SMPL Background}

The SMPL model~\cite{loper2023smpl} is a 3D surface-based human body model with $N~=~6,890$ vertices. 
The model is parameterized by a kinematic tree containing $K = 23$ joints. 
The body pose $\theta \in \mathbb{R}^{72}$ is represented as rotations at these $K$ joints in angle-axis format, plus a global rotation. 
The global translation is represented by $t \in \mathbb{R}^{3}.$
The canonical pose $\theta^*$ (T-pose) standardizes body poses across subjects. 
PCA is applied to a dataset of T-pose shapes, yielding a mean shape $\bar{\mathbf{T}}$ and shape principal components (PCs) $B_S(\beta)$ that capture shape variation over a population. 
Subject-specific canonical shape $\textbf{T}_S = \bar{\mathbf{T}} + B_S(\beta) \in \mathbb{R}^{3N}$ is computed from a shape vector $\beta \in \mathbb{R}^{10}$.
Joint locations $\mathbf{J} = \mathcal{J} \textbf{T}_S$ are linearly regressed from the canonical shape using a subject-agnostic joint regression matrix $\mathcal{J} \in \mathbb{R}^{3K \times 3N}$.
In addition, the model defines pose blend shapes~$B_P(\theta; \mathcal{P})= \left( R(\theta) - R(\theta^*) \right) \mathcal{P}$ to handle pose-induced deformations, where $R \in \mathbb{R}^{9K}$ are the vectorized rotation matrices of all joints, and $\mathcal{P} \in \mathbb{R}^{3N \times 9K}$ represents pose-induced deformations learned from data. 
$T_P(\beta, \theta) = \bar{\mathbf{T}} + B_S(\beta) + B_P(\theta)$ includes shape- and pose-induced corrections.
Each vertex $\mathbf{v_n} \in T_P(\beta, \theta)$ is transformed using linear blend skinning $\mathbf{v}_n^\prime = \sum_{k=1}^{K} w_{k,n} G_k^\prime \mathbf{v}_n$,
where~$w_{k,n}$~is the weights of joint~$k$~for vertex~$n$, and~$G_k^\prime$~is the relative transformation of joint $k$. 
We refer the readers to the SMPL paper for more details~\cite{loper2023smpl}. 
 
\subsection{Training Fetal Model}

For subject $i$ and volume $j$ in a given time series, our training data comprises keypoints $\mathbf{K}_{i,j} \in \mathbb{R}^{3 \times 16}$ (which do not necessarily map one-to-one to SMPL joints) and body meshes $\textbf{V}_{i,j} \in \mathbb{R}^{3 \times N_j}$.
We use the marching cubes algorithm~\cite{lorensen1998marching} to extract surface vertices from the volumetric segmentations.
To construct the fetal body model, we estimate the canonical shape $\mathbf{T}_S$ for each subject and the body pose $\theta$ for each frame in the time series.
Applying PCA to $\mathbf{T}_S$ across all subjects yields the mean shape $\bar{\mathbf{T}}$ and shape PCs $B_S$. 
We learn the pose blend shapes~$\mathcal{P}$ and the anatomical keypoint regressor $\mathcal{J}_{\text{k}}$ from our fetal data.

We use a coordinate descent algorithm to alternate between the estimation of the body pose and canonical shape. 
We estimate the body pose using initial canonical shapes.
Then, we ``unpose'' surface vertices to the canonical space using the most recent body pose estimate. 
Next, we refine canonical shape estimates using all unposed vertices across the entire time series. 
We iterate these steps until convergence. 
In the following sections, we describe each step in more detail.

\subsubsection{Initialization.}
We start the iterative process by registering the SMIL~model~\cite{hesse2019learning} (\ie, an infant SMPL model) to segmentation vertices $\mathbf{V}_{i,j}$ and keypoints $\mathbf{K}_{i,j}$.
We fix SMIL model's $\{\bar{\mathbf{T}}, \mathcal{S}, \mathcal{W}\}$ and optimize pose $\theta_{i,j}$, translation $t_{i,j}$, and shape parameters $\beta_i$ for each subject $i$ by minimizing
\begin{align}
E_{\text{init}} = \sum_{j} \left( E_{\text{v}} + \lambda_{\text{k}} E_{\text{k}} \right) + \lambda_{\text{shape}} E_{\text{shape}} + \lambda_{\text{smooth}} E_{\text{smooth}} + \lambda_{\text{prior}} E_{\text{prior}}.
\end{align}
The vertex term $E_{\text{v}}$ is the two-sided Chamfer distance between the model and segmentation vertices $\mathbf{V}_{i,j}$.
The keypoint term $E_{\text{k}} = \sum_{k} \rho \left( \| \mathcal{J}_{\text{k}} \mathbf{T}_{i,j} - \mathbf{K}_{i,j} \| \right)$ uses the generalized robust loss function~\cite{barron2019general} $\rho(\cdot)$ to measure distances between the model-predicted keypoints and the annotated keypoints $\mathbf{K}_{i,j}.$
Transformed model vertices are given by $\mathbf{T}_{i,j} = W(T_{S,i}+B_P(\theta_{i,j}), \mathbf{J}, \theta_{i,j}, \mathcal{W}) + {t_{i,j}}$.
$E_{\text{shape}} = \lVert \beta_i \rVert^2$ regularizes shape vectors.
We add a temporal smoothness term $E_{\text{smooth}}$ that imposes $\ell_2$ penalty on the differences in pose $\theta_{i,j}$ and translation $t_{i,j}$ between consecutive frames to encourage smooth motion. 
To ensure the pose remains kinematically feasible, we encourage the predicted poses to be close to a multivariate Gaussian distribution measured using the squared Mahalanobis distance
$E_{\text{prior}} = \sum_{j} (\theta_{i,j} - \bar{\theta})^\top \Sigma^{-1} (\theta_{i,j} - \bar{\theta})$
where $\bar{\theta}$ denotes the mean pose and $\Sigma$ represents the covariance matrix of the SMIL pose prior.
We initialize the pose blend shapes $\mathcal{P}$ to zero and the anatomical keypoint regressor $\mathcal{J}_{\text{k}}$ to be equal to the corresponding joints from the adult regressor $\mathcal{J}_{\text{A}}$.
$\lambda_{\text{k}}$, $\lambda_{\text{shape}}$, $\lambda_{\text{smooth}}$, and $\lambda_{\text{prior}}$ are the corresponding regularization coefficients.

\subsubsection{Step 1: Pose.}
We fix the subject-specific canonical shape $\mathbf{T}_{S,i}$, the pose blend shapes $\mathcal{P}$, and the anatomical keypoint regressor $\mathcal{J}_{\text{k}}$, and optimize the pose parameters $\theta_{i,j}$ and translations $t_{i,j}$ for each subject $i$ and frame $j$ to minimize: 
\begin{align}
E_{\text{pose}} = E_{\text{v}} + \lambda_{\text{k}} E_{\text{k}} + \lambda_{\text{smooth}} E_{\text{smooth}} + \lambda_{\text{prior}} E_{\text{prior}},
\end{align}
with all terms defined above in the initialization step.
We initialize the pose parameters ${\theta_{i,j}}$ and translations ${t_{i,j}}$ using the results from the previous step.
 
\subsubsection{Step 2: Pose Blend Shapes and Anatomical Keypoint Regressor.}
Next, we fix the pose parameters ${\theta_{i,j}}$ and the canonical shapes $\mathbf{T}_{S,i}$, and optimize $\mathcal{P}$ by minimizing:
\begin{align}
E_{\mathcal{P}} = \sum_{i,j} E_{\text{v}} + \lambda_{\text{P}} E_{\text{P}} + \lambda_{\text{elast}} E_{\text{elast}},
\end{align}
where $E_{\text{P}} = \lVert \mathcal{P} \rVert_F^2$ penalizes the quadratic Frobenius norm of the pose blend shapes to prevent overfitting.
The elastic regularization term $E_{\text{elast}} = \sum_{e} \lVert \left( \mathbf{T}_{S,i} \right)_e - \left( \mathbf{T}_{S,i}^{0} \right)_e \rVert^2$ penalizes changes in edge vectors between the optimized meshes and their initial value.
Here, $(\mathbf{T}_{S,i})_e$ and $(\mathbf{T}_{S,i}^{0})_e$ denote edge vectors of the optimized and initial meshes, respectively.
$\lambda_{\text{P}}$ and $\lambda_{\text{elast}}$  are corresponding weights.

Next, we update the anatomical keypoint regressor $\mathcal{J}_{\text{k}}$. 
We fix the pose parameters ${\theta_{i,j}}$, the canonical shapes $\mathbf{T}_{S,i}$, and the pose blend shapes $\mathcal{P}$, and optimize $\mathcal{J}_{\text{k}}$ by minimizing:
\begin{align}
E_{\text{key}} = \sum_{i,j} E_{\text{k}} + \lambda_{\text{J}} E_{\text{J}},
\end{align}
where $E_{\text{J}} = \lVert \mathcal{J}_{\text{k}} - \mathcal{J}_{\text{A}} \rVert^2$ is a regularization term designed to keep the anatomical keypoint regressor $\mathcal{J}_{\text{k}}$ close to the adult model regressor $\mathcal{J}_{\text{A}}.$
$\lambda_{\text{J}}$ is the corresponding weight.

\subsubsection{Step 3: Canonical Shape.}
Next, we estimate canonical shapes $\mathbf{T}_{S,i}$ using the unposed segmentation vertices $\mathbf{V}_{i,j}$. 
For each vertex $\mathbf{v} \in \mathbf{V}_{i,j}$, we find the nearest model vertex $\mathbf{u}_k \in \mathbf{T}_{i,j}$ and apply the inverse transformation to map it to the canonical space:
\begin{align}
\tilde{\mathbf{v}} = \left( G_k^\prime(\theta_{i,j}, \mathbf{J}) \right)^{-1} \left( \mathbf{v} - t_{i,j} \right).
\end{align}
We subtract the corresponding pose blend shape $B_P(\theta_{i,j})$ from the unposed vertices to isolate the canonical shape contribution. 
Then, we optimize the canonical shape $\mathbf{T}_{S,i}$:
\begin{align}
E_{\text{shape}} = \sum_{i} \left( E_{\text{Chamfer}} + \lambda_{\text{elast}} E_{\text{elast}} \right),
\end{align}
where $E_{\text{Chamfer}}$ is the two-sided Chamfer distance between the canonical shape $\mathbf{T}_{S,i}$ and the unposed surface vertices.
The elastic regularization term $E_{\text{elast}}$ is defined in Step 2 and $\lambda_{\text{elast}}$ is its weighting factor.

\subsubsection{Mean Canonical Shape $\bar{\mathbf{T}}$ and Shape Principal Components $\mathcal{S}$.}
We repeat the above pose and shape estimation steps (Steps $1$-$3$) until convergence.
Finally, we compute the mean canonical shape $\bar{\mathbf{T}}$ by averaging the optimized $\mathbf{T}_{S,i}$ across all subjects.
We then apply robust PCA~\cite{candes2011robust} to the residuals to derive the shape blend shapes~$\mathcal{S}$. 
The mean canonical shape~$\bar{\mathbf{T}}$, shape blend shapes~$\mathcal{S}$, pose blend shapes $\mathcal{P}$, and the anatomical landmark regressor $\mathcal{J}_{\text{k}}$ together constitute our fetal body model.

\subsubsection{New Subjects.}
For a new subject, we register the fetal model to segmentation vertices $\mathbf{V}_{i,j}$ and keypoint time series $\mathbf{K}_{i,j}$. 
This process mimics the model training procedure but keeps the model parameters fixed.
We initialize the pose~$\theta_{i,j}$ and the shape $\beta_{i}$ parameters following the ``Initialization'' step using the trained fetal model.
We then unpose the vertices to the canonical space.
Next, we optimize the shape parameters $\beta_{i}$ using the two-sided Chamfer distance to estimate the canonical shape $\mathbf{T}_{S}$.
Finally, we refine the pose by optimizing $\theta_{i,j}$ with updated shape $\mathbf{T}_{S}$.
Using the canonical shape $\mathbf{T}_{S}$, we assess clinically relevant anthropometric parameters, including body length (BL), head circumference (HC), abdominal circumference (AC), body volume (BV), and femur length (FL).

\subsubsection{Implementation Details.} 
We use the Adam optimizer~\cite{kingma2014adam} with a learning rate of $1e^{-3}$.
We adopt the PyTorch SMPL-X implementation of SMPL~\cite{pavlakos2019expressive}.
For efficient 3D nearest neighbor search, we use the KD-tree implementation in PyTorch3D~\cite{ravi2020pytorch3d}.
We use AltViewer~\cite{Kaufmann_Vechev_aitviewer_2022} to visualize the meshes and keypoints.

%% file: texts/experiment.tex
\section{Experiments and Results}
\label{sec:experiment}

\subsubsection{Data and Evaluation.}
Our dataset comprises two cohorts: a research cohort and a clinical cohort, both from Boston Children's Hospital. 
The research cohort includes $53$ singleton subjects with gestational age (GA) of $24$ -- $37$ weeks, imaged with EPI time-series of 216 $\pm 125$ volumetric frames.
The clinical cohort consists of $11$ subjects, including $6$ diagnosed with Chiari II malformation and $5$ non-Chiari controls. 
The median (IQR) of GA among the clinical cohort is $24$wk ($22$-$35$wk).
All scans were acquired on a $3$T Siemens Skyra scanner.

To test the model's generalization, we created the ``New Shape'' and ``New Pose'' test sets. 
We randomly selected $20\%$ subjects from the research cohort to form the ``New Shape'' subset. 
For the remaining subjects, we leave out 10\% of the frames in each time series to form the ``New Pose'' dataset. 
The ``New Pose'' dataset evaluates generalization to unseen poses, and the ``New Shape'' dataset assesses generalization to unseen shapes.
We repeat the above data selection process to obtain five non-overlapping ``New Shape'' test sets.
This ensures each subject contributes once to the ``New Shape'' dataset.

We curated a ground truth fetal body segmentation dataset and trained a UNet-based model to obtain fetal body segmentations~\cite{billot2024network,liu2023consistency}.
Keypoints were obtained using a pre-trained model~\cite{xu20203d,xu2019fetal,zhang2020enhanced}.
The trained fetal body model was then registered to the clinical cohort to measure anthropometric measurements.
To quantitatively evaluate the alignment, we compute the one-sided median Chamfer distance from the segmentation surface to the model. 
To derive a metric for time series, we average the median Chamfer distances over time. 

\input{figtexs/iter_n_beta_baseline_miccai.tex}

\subsubsection{Baselines.} 
While our closest baseline model is infant SMPL~\cite{hesse2019learning}, it may not effectively capture fetal body shape since newborns are typically larger than fetuses and their bodies have different proportions.
We downscale the newborn model to $75\%$ of its size to match the fetal population~\cite{hesse2019learning,snijders1994fetal}.
We evaluate two variants of the infant SMPL model. 
The first variant --- infant SMPL --- jointly estimates $\beta$ and $\theta$ without an unpose step.
The second variant --- infant SMPL w. unpose --- decouples the estimation of $\beta$ and $\theta$ with an unposing step, following our registration method. 

\subsubsection{Quantitative Results.}
\figref{fig:iter_n_beta_baseline}A reports the alignment error as a function of training iterations. 
The model achieved a median alignment error of $3.01$~mm on the training dataset in the third iteration, after which the performance plateaued. 
Similarly, in the ``New Pose'' and ``New Shape'' datasets, the model achieves errors of $3.14$~mm and $3.17$~mm. 
These results indicate that our model generalizes well to unseen poses and shapes. 
We use the $3$rd iteration model in the following experiments.
\figref{fig:iter_n_beta_baseline}B reports the alignment error as a function of number of principal components (PCs).
The alignment error begins to converge at approximately 8 PCs.
Interestingly, while including more PCs enables finer shape adjustments, it results in a qualitatively less smooth body surface.
To balance the quality of fit and that of visualization, we use the first $10$ PCs, which capture $98.3\%$ of the variance, consistently with SMPL and SMIL.

\figref{fig:iter_n_beta_baseline}C compares our method with the two baseline methods on the ``New Shape'' dataset. 
Across all subjects, infant SMPL model achieves a median alignment error of $5.21$~mm.
We observe that if the shape parameters $\beta$ and pose parameters $\theta$ are estimated jointly, the predicted fetal body shapes are typically smaller (infant SMPL).
This is likely because in the fetal pose, the body parts are clumped together.
A smaller predicted body shape results in lower alignment error but does not reflect the true body shape.
If we decouple the shape and pose estimation using the ``unpose'' step, the body alignment improves significantly to $3.61$~mm (infant SMPL w. unpose).
Finally, our model with fully decoupled estimation achieved the lowest alignment error of $3.07$~mm. 

We train the final model using the entire research cohort to fully utilize our data.
We then align the model to all research subjects to evaluate clinically relevant anthropometric measurements.
The model-derived metrics are strongly correlated with gestational age (GA) (HC: $R=0.79,p<0.001$, AC: $R=0.80,p<0.001$, FL: $R=0.77,p<0.001$, BL: $R=0.78,p<0.001$, BV: $R=0.82,p<0.001$) consistent with prior literature~\cite{snijders1994fetal}.
We align the model to the clinical cohort, and the median segmentation-to-model Chamfer distance is $4.2$~mm.

\input{figtexs/shape_magnification_and_pc.tex}

\subsubsection{Visualization.}
In \figref{fig:eigen_shape_and_shape_magnification}A, we visualize population-level shape characteristics using the learned eigen-shapes.
The first three PCs respectively capture body size, abdominal thickness, and ratio of leg length to body length.
In \figref{fig:eigen_shape_and_shape_magnification}C, we visualize and compare canonical shapes from three subjects ($\text{S0}$, $\text{S1}$, and $\text{S2}$).
We choose $\text{S0}$ to be the one closest to the mean shape.
We select $\text{S1}$ and $\text{S2}$ based on their deviations in the second principal component. 
We ensure they have small deviations in the first PC to isolate the effects of the second PC.
Shape differences among $\text{S1}$, $\text{S2}$, and $\text{S0}$ are visually subtle (\figref{fig:eigen_shape_and_shape_magnification}D).
To amplify these differences, we magnify the deviations of $\text{S1}$ and $\text{S2}$ relative to $\text{S0}$ by a factor of two (\figref{fig:eigen_shape_and_shape_magnification}D).
The resulting magnified $S1+$ and $S2+$ shapes show clear differences in body thickness compared to $\text{S0}$.
Our tool enables the visualization and amplification of subtle shape differences among subjects for improved comparison.

%% file: figtexs/iter_n_beta_baseline_miccai.tex
\begin{figure*}[t]
    \centering
    \includegraphics[width=\linewidth]{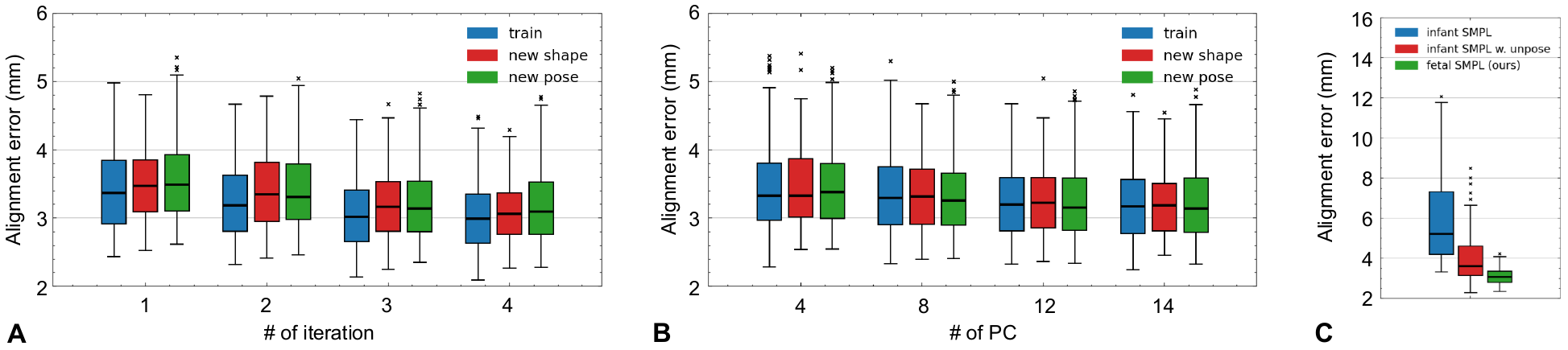}
    \caption{
        \textbf{Alignment error statistics.}
        (\textbf{A}) Alignment error decreases as we train the model for more iterations., and plateaus after $3$ iterations. Model generalizes well to new shapes and new poses. 
        (\textbf{B}) Alignment error declines as we use more PCs. The improvement after $12$th PC is marginal. 
        (\textbf{C}) Fetal model ourperforms baseline models.
    }
    \label{fig:iter_n_beta_baseline}
\end{figure*}

%% file: figtexs/shape_magnification_and_pc.tex
\begin{figure}[t]
    \centering
    \includegraphics[width=\linewidth]{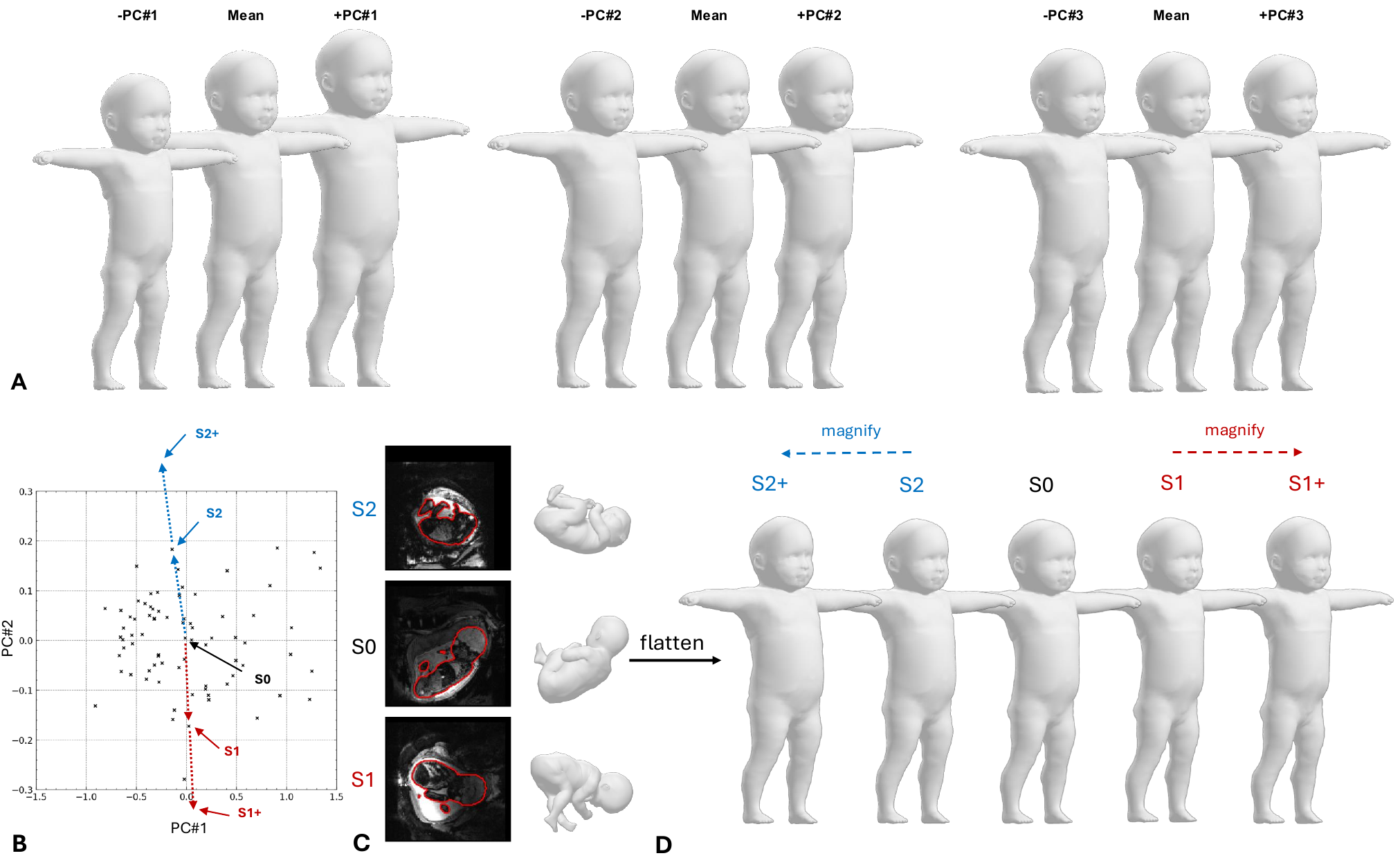}
    \caption{
        \textbf{Inter-subject shape differences and population-level shape PCs.}
        (\textbf{A}) The first $3$ PCs describe body size, body thickness, and ratio between leg length and body length. PC1 varies from $-2$ to $+2$ standard deviations from left to right, while PC2 and PC3 vary from $-4$ to $+4$ to make shape differences more visible.
        (\textbf{B}) Scatter plot of the first two PCs with $\text{S0}$, $\text{S1}$, and $\text{S2}$ subjects marked in the distribution.
        (\textbf{C}) The MR images and posed body shapes of $\text{S0}$, $\text{S1}$, and $\text{S2}$.
        (\textbf{D}) Side by side visualization of $\text{S2+}$, $\text{S2}$, $\text{S0}$, $\text{S1}$, and $\text{S1+}$. 
    }
    \label{fig:eigen_shape_and_shape_magnification}
\end{figure}

%% file: texts/conclusion.tex
\section{Conclusions}
\label{sec:conclusion}

In this paper, we presented a 3D articulated statistical fetal body model.
We trained the model from noisy segmentations and keypoints extracted from the MRI time series. 
A key enabler is alternating between the estimation of the shape and pose parameter.
The resulting model accurately explains unseen fetal shapes.
Moreover, the model provides clinically relevant measurements that are traditionally difficult to obtain from segmentations and keypoints.
This work promises to provide improved fetal development assessment and better prenatal care.

\subsubsection{Acknowledgements.}
This work has been funded by NIH NIBIB 1R01EB032708, NIH NICHD 1R01HD114338, NIH NIBIB 1R01EB036945, and MIT CSAIL-Wistron Program.
We are grateful to Neerav Karani, Neel Dey, Marilyn Keller, Nikolas Hesse, Sergi Pujades, S. Mazdak Abulnaga, and Justin Solomon for their insightful discussions. 
We also extend our thanks to the MIT Medical Vision Group and the Fetal Neonatal Neuroimaging and Developmental Science Center at Boston Children's Hospital for their invaluable support.

\subsubsection{Disclosure of Interests.}
The authors have no competing interests to declare.

%% file: texts/appendix.tex
\section{Method}
\label{sec:supp_method}

\subsection{Anthropometric Measurements}

We provide a detailed description of the method for computing anthropometric measurements from the 3D fetal body model. 
All measurements are derived from the canonical shape mesh.

\vspace{0.5em}
\noindent \textbf{Body Length (BL)}: Defined as the distance between the vertex at the top of the head and the midpoint of the two foot-bottom vertices.

\vspace{0.5em}
\noindent \textbf{Femur Length (FL)}: Calculated as the distance between the hip and knee anatomical keypoints. 
We report the average of the left and right femur lengths.

\vspace{0.5em}
\noindent \textbf{Body Volume (BV)}: Computed using the divergence theorem. 
We define a vector field $\mathbf{F}(\mathbf{r}) = \frac{1}{3} \mathbf{r}$, where $\mathbf{r}$ is the position vector. 
The divergence of $\mathbf{F}$ is constant and equal to $1$ everywhere in space. 
The volume is then calculated as the flux of $\mathbf{F}$ through the body surface mesh: 
\begin{align}
V = \int_V 1 \, dV = \int_V (\nabla \cdot \mathbf{F}) \, dV = \oint_S \mathbf{F} \cdot \mathbf{n} \, dS, 
\end{align}
where $\mathbf{n}$ denotes the outward normal vector on the surface $S$.

\vspace{0.5em}
\noindent \textbf{Head Circumference (HC) and Abdominal Circumference (AC)}: 
We slice the canonical shape mesh with planes defined by anatomical landmarks. 
The intersection yields line segments, from which we extract unique intersection points.
The circumference is then calculated as the perimeter of this hull.

\subsection{Training Hyperparameters} We report the hyperparameters used in training the model.

\begin{itemize}
    \item \textbf{Init.:} $LR=0.003$, $\lambda_{\text{k}} = 1.0$, $\lambda_{\text{smooth}} = 0.001$, $\lambda_{\text{prior}} = 0.1$, $\lambda_{\text{shape}} = 0.1$.
	\vspace{0.2em}
    \item \textbf{Step $1$:} $LR=0.003$, $\lambda_{\text{k}} = 1.0$, $\lambda_{\text{smooth}} = 0.001$, $\lambda_{\text{prior}} = 0.1$.
   	\vspace{0.2em}
    \item \textbf{Step $2$:} $LR=0.003$, $\lambda_{\text{elast}} = 5$, $\lambda_{\text{P}} = 0.01$, $\lambda_{\text{J}} = 0.01$.
   	\vspace{0.2em}
    \item \textbf{Step $3$:} $LR=0.005$, $\lambda_{\text{elast}} = 5$.
\end{itemize}

\subsection{Generalized Robust Loss Function}

We employ the generalized robust loss function~\cite{barron2019general} to measure the difference between the predicted and ground truth segmentation vertex and keypoints:
\begin{align}
\rho(x; \alpha, c) = \frac{|\alpha - 2|}{\alpha} \left( \left( \frac{x}{c} \right)^2 / |\alpha - 2| + 1 \right)^{\alpha/2} - 1,
\end{align}
where $\alpha \in \mathbb{R}$ controls the robustness and $c > 0$ controls scale. 
We used $\alpha = -1.5$ and $c = 1$ in our experiments.

\section{Additional Results}

\subsection{Research Cohort}

\input{figtexs/pc_and_subj_additional_results.tex}

\paragraph{Body Shape Visualization} 

In the main paper, we selected $\text{S0}$, $\text{S1}$, and $\text{S2}$ using the first two shape PCs. 
Here, we instead use the first and third PCs to select $\text{S3}$ and $\text{S4}$.
\figref{fig:subj_shape_diff_addi_results}A shows a scatter plot of the first and third PCs with $\text{S0}$, $\text{S3}$, and $\text{S4}$ marked in the distribution.
To highlight shape differences, we extrapolate \text{S3} and \text{S4}. 
We define $\text{S3+}$ and $\text{S4+}$ as the shapes that are twice the distance from $\text{S0}$ in the direction of $\text{S3}$ and $\text{S4}$, respectively.
\figref{fig:subj_shape_diff_addi_results}B visualizes $\text{S3}$, $\text{S0}$, and $\text{S4}$ with their corresponding MR images and posed body meshes.
\figref{fig:subj_shape_diff_addi_results}C flattens the body shape and shows the side-by-side visualization of $\text{S3+}$, $\text{S3}$, $\text{S0}$, $\text{S4}$, and $\text{S4+}$.
While the shape differences between $\text{S3}$, $\text{S0}$, and $\text{S4}$ are subtle, those between $\text{S3+}$, $\text{S0}$, and $\text{S4+}$ are more visually pronunced.

\paragraph{Alignment Error} 

We present additional alignment results as a function of training iterations and the number of shape PCs.
In the main paper, we report the segmentation-to-model Chamfer distance. 
Here we provide model-to-segmentation Chamfer distance and L2 error of all anatomical keypoints (\ie, ankle, bladder, knee, eye, elbow, shoulder, hip, neck, and wrist).
\figref{fig:error_iter_all} shows decreasing alignment error with more iterations.
\figref{fig:error_num_beta_all} shows lower alignment error as a function of number of PCs in the model.

\paragraph{Anthropometric Measurements}

\input{figtexs/anthropometric_measurements.tex}

In the main paper, we report the correlations between the model-derived anthropometric measurements and gestational age (GA).
Here, we present the corresponding scatter plots in \figref{fig:anthropometric_measurements}.

\subsection{Clinical Cohort}

\input{figtexs/clinical_alignment.tex}
\input{figtexs/anthro_all_boxplots.tex}

\paragraph{Alignment Error} 

\figref{fig:clinical_alignment} shows alignment results for the clinical cohort.
The median model-to-segmentation Chamfer distance is $3.9$mm (IQR: $3.6$-$4.7$mm).
The segmentation-to-model alignment error is higher ($5.9$mm).
This is in part due to the false positive segmentations in the clinical images (\eg, umbilical cord, placenta, and maternal body).
These false positive segmentations bias segmentation-to-model alignment towards higher error.
However, they typically do not affect the model-to-segmentation alignment evaluation because the nearest segmentation surface vertex of the model is often close to the true fetal body surface.
For keypoints, our model achieved good results. 
Median L2 errors are lower than $6.0$mm for all keypoints.

\paragraph{Anthropometric Measurements}

\figref{fig:anthro_all_boxplots} shows the anthropometric measurements of the clinical cohort and the research cohort.
The clinical Chiari cohort has a similar gestational age distribution as the research cohort.
However, the clinical control cohort shows generally lower measurements, likely due to the younger GA.
A large sample size and rigorous statistical testing are needed to assess the significance.

\input{figtexs/error_vs_iter_all.tex}

\input{figtexs/error_vs_num_beta_all.tex}

%% file: figtexs/pc_and_subj_additional_results.tex
\begin{figure*}[t]
    \centering
    \includegraphics[width=\linewidth]{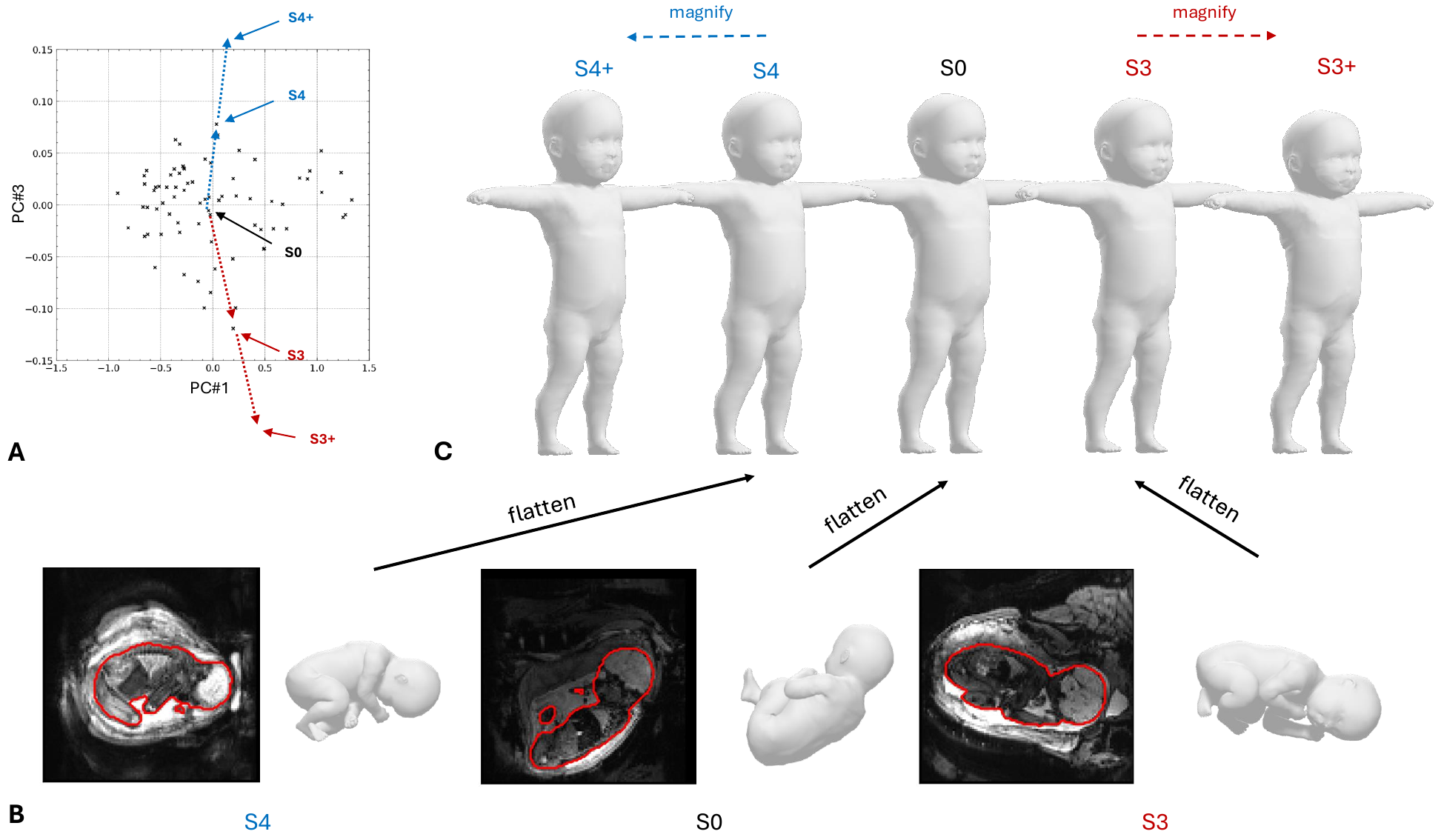}
    \caption{
        \textbf{Inter-subject shape differences (additional results)}.
        (\textbf{A}) Scatter plot of the first and third PCs with $\text{S0}$, $\text{S3}$, and $\text{S4}$ subjects marked in the distribution.
        (\textbf{B}) The MR images and posed body shapes of $\text{S0}$, $\text{S3}$, and $\text{S4}$.
        (\textbf{C}) Side by side visualization of $\text{S4+}$, $\text{S4}$, $\text{S0}$, $\text{S3}$, and $\text{S3+}$.
    }
    \label{fig:subj_shape_diff_addi_results}
\end{figure*}

%% file: figtexs/anthropometric_measurements.tex
\begin{figure*}[h]
    \centering
    \includegraphics[width=\linewidth]{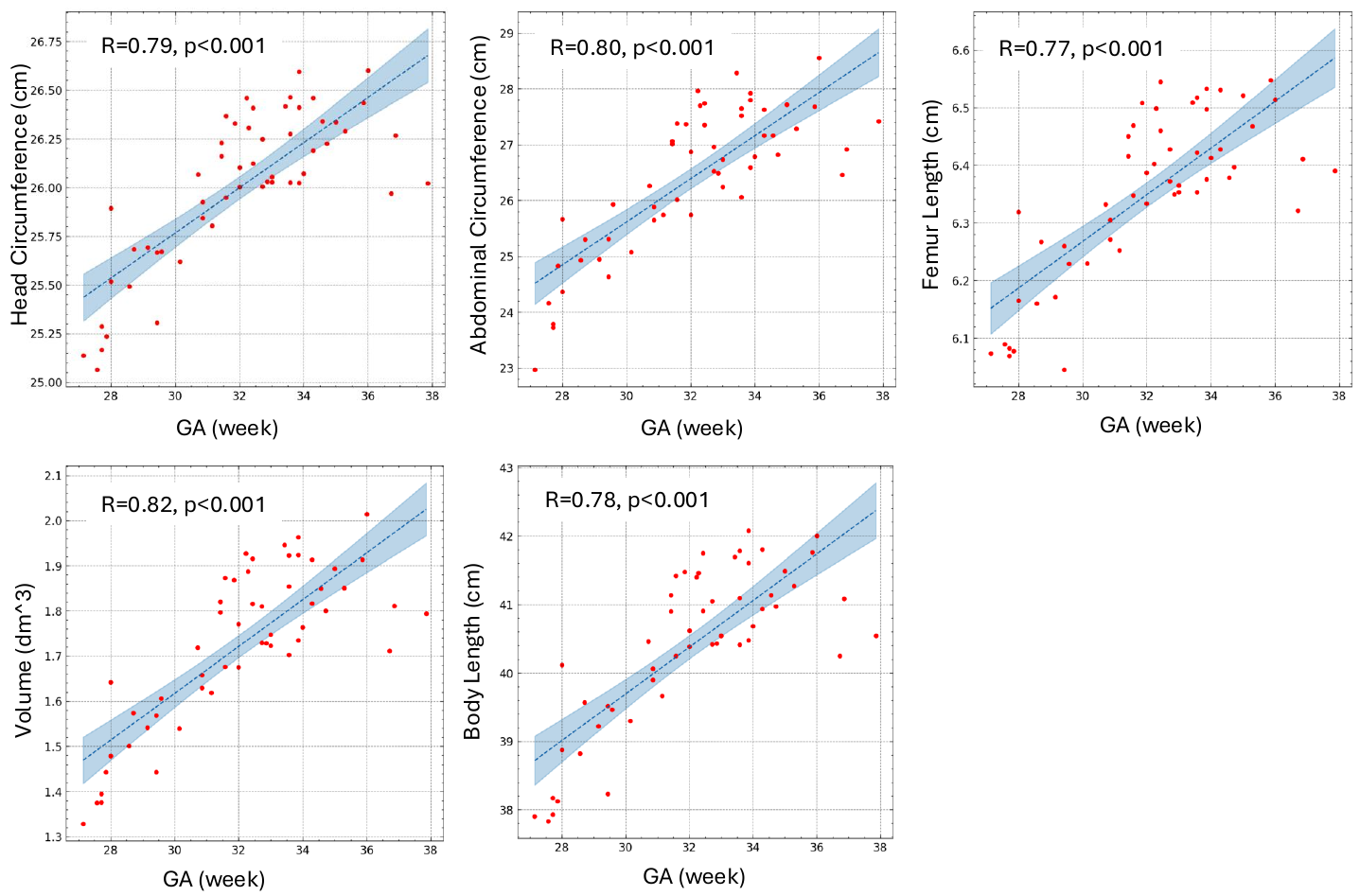}
    \caption{
        \textbf{Model-derived anthropometric measurements and GA.}
        We compare GA against model-derived FL, BL, HC, AC, and BV in the research cohort.
    }
    \label{fig:anthropometric_measurements}
\end{figure*}

%% file: figtexs/clinical_alignment.tex
\begin{figure*}[t]
    \centering
    \includegraphics[width=\linewidth]{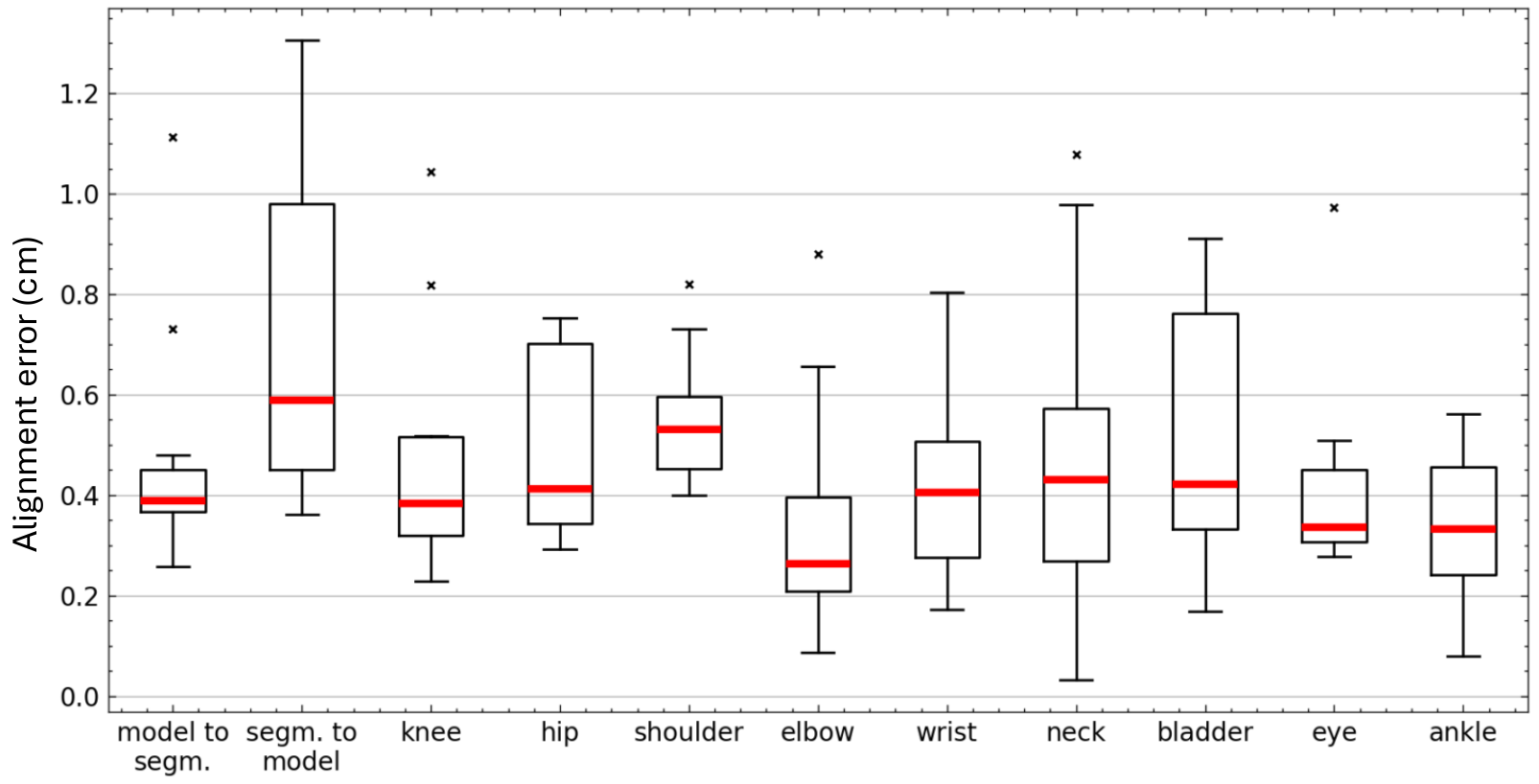}
    \caption{
        \textbf{Alignment error in clinical cohort.}
    }
    \label{fig:clinical_alignment}
\end{figure*}

%% file: figtexs/anthro_all_boxplots.tex
\begin{figure*}[h]
    \centering
    \includegraphics[width=\linewidth]{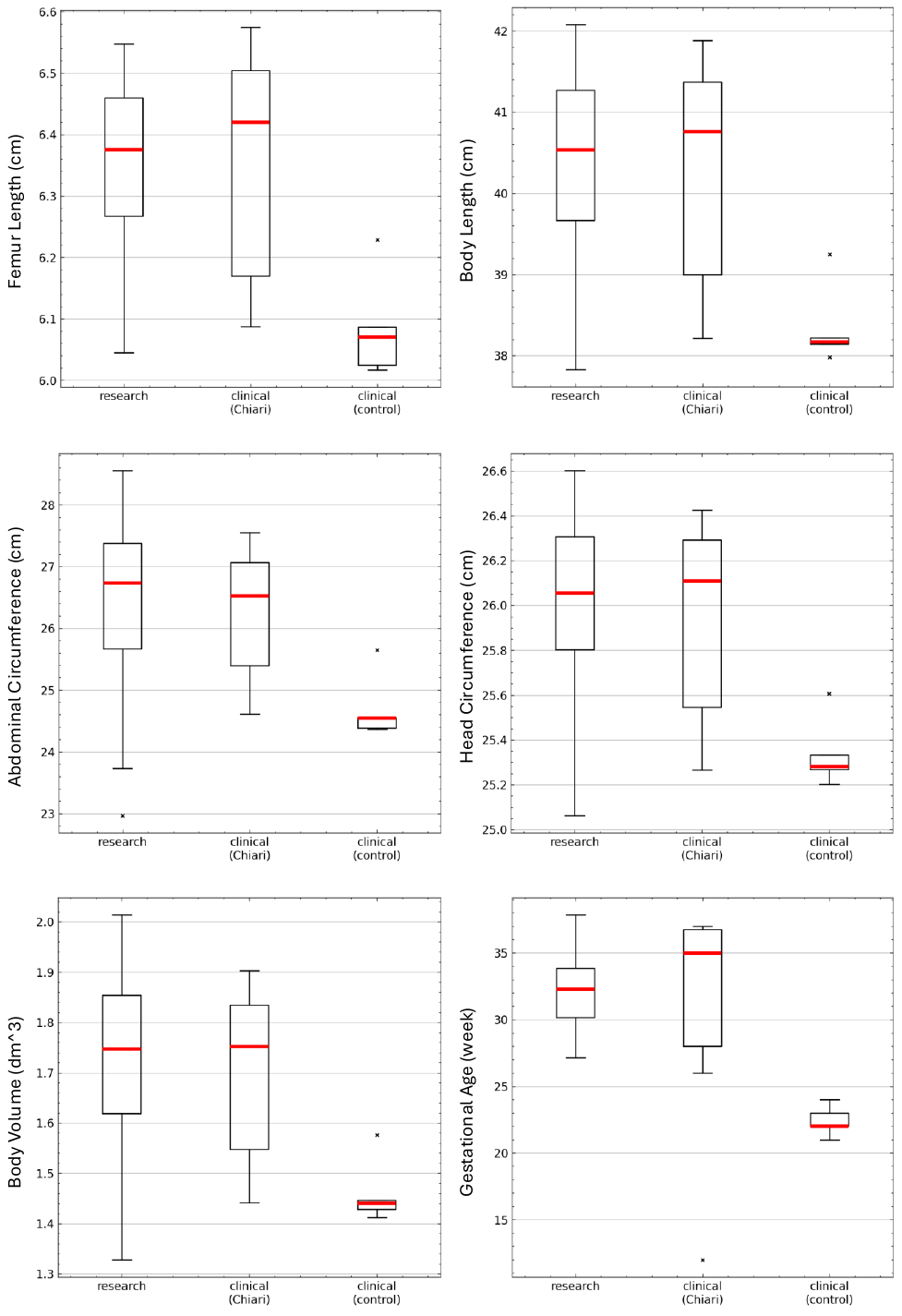}
    \caption{
        \textbf{Anthropometric measurements and gestational age in research and clinical cohort.} 
        We compare FL, BL, HC, AC, BV, and GA among research, clinical (Chiari), and clinical (control) cohorts.
    }
    \label{fig:anthro_all_boxplots}
\end{figure*}

%% file: figtexs/error_vs_iter_all.tex
\begin{figure*}[h]
    \centering
    \includegraphics[width=\linewidth]{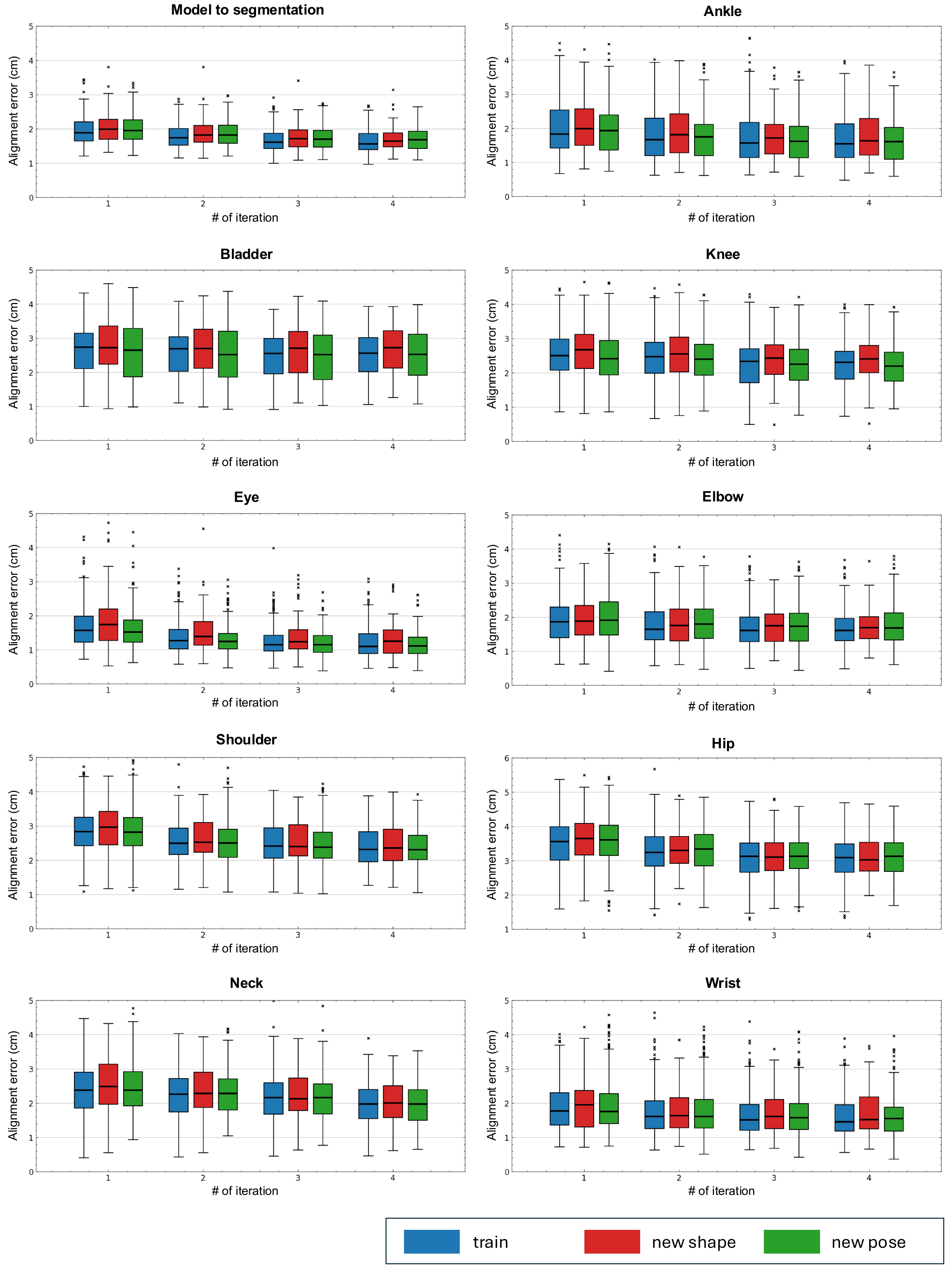}
    \caption{
        \textbf{Alignment error as a function of iterations.} Model to segmentation Chamfer distance and L2 error of all keypoints are reported.
    }
    \label{fig:error_iter_all}
\end{figure*}

%% file: figtexs/error_vs_num_beta_all.tex
\begin{figure*}[h]
    \centering
    \includegraphics[width=\linewidth]{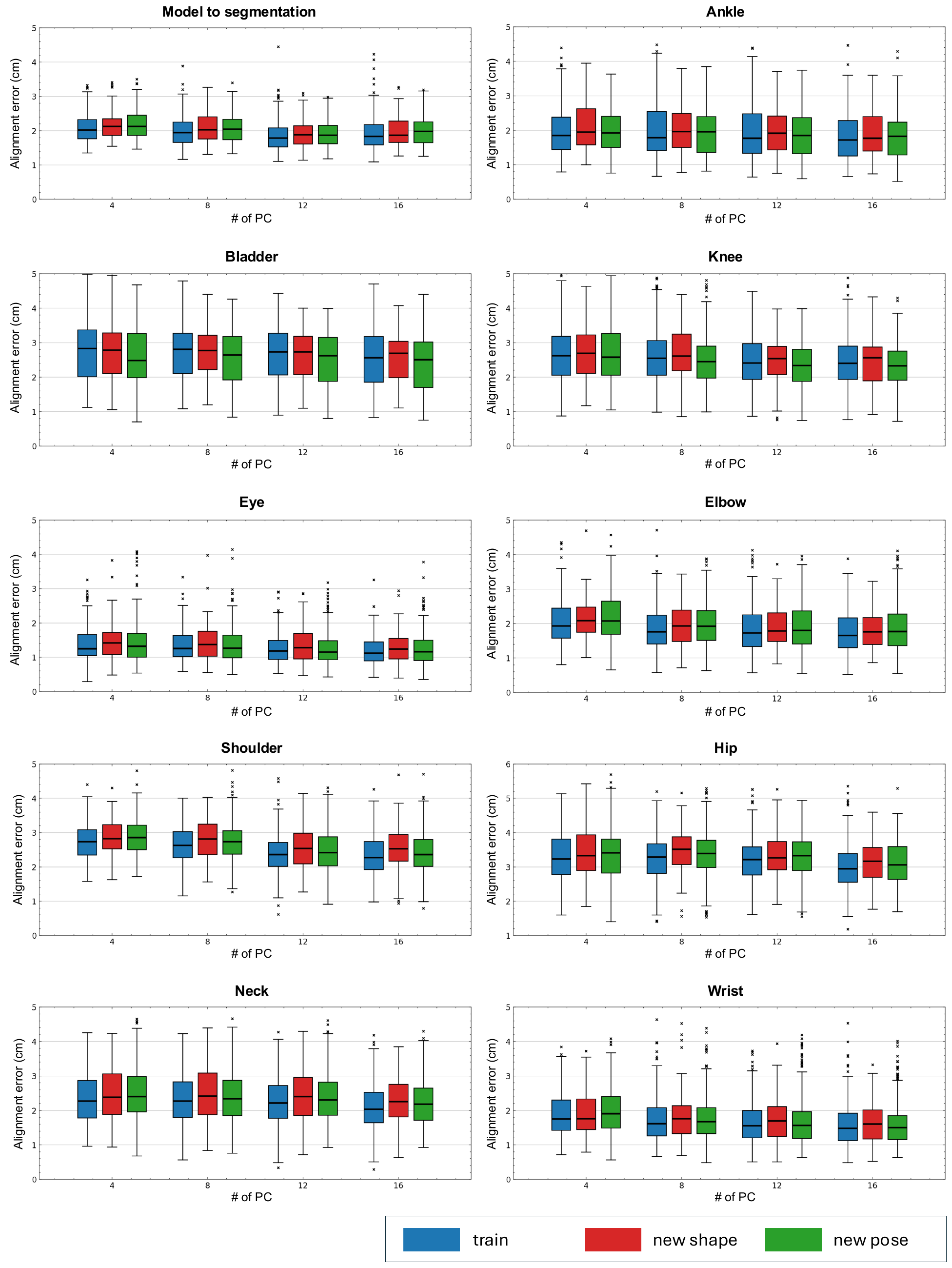}
    \caption{
        \textbf{Alignment error as a function of number of principal components.} Model to segmentation Chamfer distance and L2 error of all keypoints are reported.
    }
    \label{fig:error_num_beta_all}
\end{figure*}

%% file: main.bbl
\begin{thebibliography}{10}
\providecommand{\url}[1]{\texttt{#1}}
\providecommand{\urlprefix}{URL }
\providecommand{\doi}[1]{https://doi.org/#1}

\bibitem{alldieck2018video}
Alldieck, T., Magnor, M., Xu, W., Theobalt, C., Pons-Moll, G.: Video based reconstruction of 3d people models. In: Proceedings of the IEEE Conference on Computer Vision and Pattern Recognition. pp. 8387--8397 (2018)

\bibitem{barron2019general}
Barron, J.T.: A general and adaptive robust loss function. In: Proceedings of the IEEE/CVF conference on computer vision and pattern recognition. pp. 4331--4339 (2019)

\bibitem{billot2024se}
Billot, B., Dey, N., Moyer, D., Hoffmann, M., Turk, E.A., Gagoski, B., Grant, P.E., Golland, P.: {SE(3)}-equivariant and noise-invariant {3D} rigid motion tracking in brain {MRI}. IEEE Transactions on Medical Imaging  (2024)

\bibitem{billot2024network}
Billot, B., Dey, N., Turk, E.A., Grant, E., Golland, P.: Network conditioning for synergistic learning on partial annotations. In: Medical Imaging with Deep Learning (2024)

\bibitem{candes2011robust}
Cand{\`e}s, E.J., Li, X., Ma, Y., Wright, J.: Robust principal component analysis? Journal of the ACM (JACM)  \textbf{58}(3),  1--37 (2011)

\bibitem{chi2023dynamic}
Chi, Z., Cong, Z., Wang, C.J., Liu, Y., Turk, E.A., Grant, P.E., Abulnaga, S.M., Golland, P., Dey, N.: Dynamic neural fields for learning atlases of 4d fetal mri time-series. arXiv preprint arXiv:2311.02874  (2023)

\bibitem{hesse2019learning}
Hesse, N., Pujades, S., Black, M.J., Arens, M., Hofmann, U.G., Schroeder, A.S.: Learning and tracking the 3d body shape of freely moving infants from rgb-d sequences. IEEE transactions on pattern analysis and machine intelligence  \textbf{42}(10),  2540--2551 (2019)

\bibitem{hesse2018learning}
Hesse, N., Pujades, S., Romero, J., Black, M.J., Bodensteiner, C., Arens, M., Hofmann, U.G., Tacke, U., Hadders-Algra, M., Weinberger, R., et~al.: Learning an infant body model from rgb-d data for accurate full body motion analysis. In: Medical Image Computing and Computer Assisted Intervention--MICCAI 2018: 21st International Conference, Granada, Spain, September 16-20, 2018, Proceedings, Part I. pp. 792--800. Springer (2018)

\bibitem{Kaufmann_Vechev_aitviewer_2022}
Kaufmann, M., Vechev, V., Mylonopoulos, D.: {aitviewer} (7 2022). \doi{10.5281/zenodo.10013305}, \url{https://github.com/eth-ait/aitviewer}

\bibitem{kingma2014adam}
Kingma, D.P.: Adam: A method for stochastic optimization. arXiv preprint arXiv:1412.6980  (2014)

\bibitem{liu2023consistency}
Liu, Y., Karani, N., Abulnaga, S.M., Xu, J., Grant, P.E., Abaci~Turk, E., Golland, P.: Consistency regularization improves placenta segmentation in fetal epi mri time series. In: International Workshop on Preterm, Perinatal and Paediatric Image Analysis. pp. 77--87. Springer (2023)

\bibitem{loper2023smpl}
Loper, M., Mahmood, N., Romero, J., Pons-Moll, G., Black, M.J.: Smpl: A skinned multi-person linear model. In: Seminal Graphics Papers: Pushing the Boundaries, Volume 2, pp. 851--866 (2023)

\bibitem{lorensen1998marching}
Lorensen, W.E., Cline, H.E.: Marching cubes: A high resolution 3d surface construction algorithm. In: Seminal graphics: pioneering efforts that shaped the field, pp. 347--353 (1998)

\bibitem{pavlakos2019expressive}
Pavlakos, G., Choutas, V., Ghorbani, N., Bolkart, T., Osman, A.A., Tzionas, D., Black, M.J.: Expressive body capture: 3d hands, face, and body from a single image. In: Proceedings of the IEEE/CVF conference on computer vision and pattern recognition. pp. 10975--10985 (2019)

\bibitem{ravi2020pytorch3d}
Ravi, N., Reizenstein, J., Novotny, D., Gordon, T., Lo, W.Y., Johnson, J., Gkioxari, G.: Accelerating 3d deep learning with pytorch3d. arXiv:2007.08501  (2020)

\bibitem{sadhwani2022fetal}
Sadhwani, A., Wypij, D., Rofeberg, V., Gholipour, A., Mittleman, M., Rohde, J., Velasco-Annis, C., Calderon, J., Friedman, K.G., Tworetzky, W., et~al.: Fetal brain volume predicts neurodevelopment in congenital heart disease. Circulation  \textbf{145}(15),  1108--1119 (2022)

\bibitem{sajapala20174d}
Sajapala, S., AboEllail, M.A.M., Kanenishi, K., Mori, N., Marumo, G., Hata, T.: 4d ultrasound study of fetal movement early in the second trimester of pregnancy. Journal of Perinatal Medicine  \textbf{45}(6),  737--743 (2017)

\bibitem{snijders1994fetal}
Snijders, R., Nicolaides, K.: Fetal biometry at 14--40 weeks' gestation. Ultrasound in Obstetrics and Gynecology: The Official Journal of the International Society of Ultrasound in Obstetrics and Gynecology  \textbf{4}(1),  34--48 (1994)

\bibitem{uus2024automated}
Uus, A.U., Hall, M., Grigorescu, I., Avena~Zampieri, C., Egloff~Collado, A., Payette, K., Matthew, J., Kyriakopoulou, V., Hajnal, J.V., Hutter, J., et~al.: Automated body organ segmentation, volumetry and population-averaged atlas for 3d motion-corrected t2-weighted fetal body mri. Scientific Reports  \textbf{14}(1), ~6637 (2024)

\bibitem{vasung2023cross}
Vasung, L., Xu, J., Abaci-Turk, E., Zhou, C., Holland, E., Barth, W.H., Barnewolt, C., Connolly, S., Estroff, J., Golland, P., et~al.: Cross-sectional observational study of typical in utero fetal movements using machine learning. Developmental neuroscience  \textbf{45}(3),  105--114 (2023)

\bibitem{xu2021motion}
Xu, J., Turk, E.A., Gagoski, B., Golland, P., Grant, P.E., Adalsteinsson, E.: Motion analysis in fetal mri using deep pose estimator. In: Proceedings of the International Society for Magnetic Resonance in Medicine... Scientific Meeting and Exhibition. International Society for Magnetic Resonance in Medicine. Scientific Meeting and Exhibition. vol.~29. NIH Public Access (2021)

\bibitem{xu20203d}
Xu, J., Zhang, M., Turk, E.A., Grant, P.E., Golland, P., Adalsteinsson, E.: 3d fetal pose estimation with adaptive variance and conditional generative adversarial network. In: Medical Ultrasound, and Preterm, Perinatal and Paediatric Image Analysis: First International Workshop, ASMUS 2020, and 5th International Workshop, PIPPI 2020, Held in Conjunction with MICCAI 2020, Lima, Peru, October 4-8, 2020, Proceedings 1. pp. 201--210. Springer (2020)

\bibitem{xu2019fetal}
Xu, J., Zhang, M., Turk, E.A., Zhang, L., Grant, P.E., Ying, K., Golland, P., Adalsteinsson, E.: Fetal pose estimation in volumetric mri using a 3d convolution neural network. In: Medical Image Computing and Computer Assisted Intervention--MICCAI 2019: 22nd International Conference, Shenzhen, China, October 13--17, 2019, Proceedings, Part IV 22. pp. 403--410. Springer (2019)

\bibitem{zhang2020enhanced}
Zhang, M., Xu, J., Abaci~Turk, E., Grant, P.E., Golland, P., Adalsteinsson, E.: Enhanced detection of fetal pose in 3d mri by deep reinforcement learning with physical structure priors on anatomy. In: Medical Image Computing and Computer Assisted Intervention--MICCAI 2020: 23rd International Conference, Lima, Peru, October 4--8, 2020, Proceedings, Part VI 23. pp. 396--405. Springer (2020)

\end{thebibliography}
